\def\year{2020}\relax
\documentclass[letterpaper]{article} % DO NOT CHANGE THIS
\usepackage{aaai20}  % DO NOT CHANGE THIS
\usepackage{times}  % DO NOT CHANGE THIS
\usepackage{helvet} % DO NOT CHANGE THIS
\usepackage{courier}  % DO NOT CHANGE THIS
\usepackage[hyphens]{url}  % DO NOT CHANGE THIS
\usepackage{graphicx} % DO NOT CHANGE THIS
\usepackage{graphicx}  % DO NOT CHANGE THIS
\usepackage{multirow}
\usepackage{amssymb}
\usepackage{amsmath}

\newcommand{\namecite}[1]{\citeauthor{#1}~\shortcite{#1}}

\newcommand{\x}{{\mathbf x}}
\newcommand{\B}{{\mathbf B}}
\newcommand{\mat}[1]{{\mathbf #1}}

\newcommand{\figer}{{FIGER}}
\newcommand{\bbn}{{BBN}}
\newcommand{\ontonotesFine}{OntoNotes$_{\text{fine}}$}

\newcommand{\accuracy}{Acc.}
\newcommand{\fOneMa}{$F1_{\text{ma}}$}
\newcommand{\fOneMi}{$F1_{\text{mi}}$}

\usepackage{colortbl}
\definecolor{lightgrey}{rgb}{0.85,0.85,0.85}

\title{Improving Entity Linking by Modeling Latent \\Entity Type Information}
\author{Shuang Chen,\textsuperscript{\rm 1}\thanks{Contribution during internship at Microsoft Research.} 
    Jinpeng Wang,\textsuperscript{\rm 2} 
    Feng Jiang,\textsuperscript{\rm 1,3}
    Chin-Yew Lin\textsuperscript{\rm 2} \\
    \textsuperscript{\rm 1} Harbin Institute of Technology, Harbin, China \\
    \textsuperscript{\rm 2} Microsoft Research Asia ~~~~\
    \textsuperscript{\rm 3} Peng Cheng Laboratory \\
    hitercs@gmail.com, \{jinpwa, cyl\}@microsoft.com, fjiang@hit.edu.cn
}
 
\begin{document}

\maketitle

\begin{abstract}
    Existing state of the art neural entity linking models employ attention-based bag-of-words context model and pre-trained entity embeddings bootstrapped from word embeddings to assess topic level context compatibility. However, the latent entity type information in the immediate context of the mention is neglected, which causes the models often link mentions to incorrect entities with incorrect type. To tackle this problem, we propose to inject latent entity type information into the entity embeddings based on pre-trained BERT. In addition, we integrate a BERT-based entity similarity score into the local context model of a state-of-the-art model to better capture latent entity type information. Our model significantly outperforms the state-of-the-art entity linking models on standard benchmark (AIDA-CoNLL). Detailed experiment analysis demonstrates that our model corrects most of the type errors produced by the direct baseline.
\end{abstract}

\section{Introduction}
Entity Linking (EL) is the task of disambiguating textual mentions to their corresponding entities in a reference knowledge base (e.g., Wikipedia). An accurate entity linking system is crucial for many knowledge related tasks such as question answering~\cite{yih-etal-2015-semantic} and information extraction~\cite{hoffmann-etal-2011-knowledge}.

Traditional entity linkers mainly depend on manually designed features to evaluate the local context compatibility and document-level global coherence of referent entities~\cite{cheng-roth-2013-relational,durrett-klein-2014-joint}. 
The design of such features requires entity-specific domain knowledge. These features can not fully capture relevant statistical dependencies and interactions. 
One recent notable work~\cite{ganea2017deep} instead pioneers to rely on pre-trained entity embeddings, learnable context representation and differentiable joint inference stage to learn basic features and their combinations from scratch. Such model design allows to learn useful regularities in an end-to-end fashion and eliminates the need for extensive feature engineering.
It also substantially outperforms the traditional methods on standard benchmark (e.g., AIDA-CoNLL).
A line of follow-up work \cite{le-titov-2018-improving,le-titov-2019-boosting,le-titov-2019-distant} investigate potential improvement solution or other task settings based on that.

\begin{figure}
    \centering
    \includegraphics[width=0.90\columnwidth]{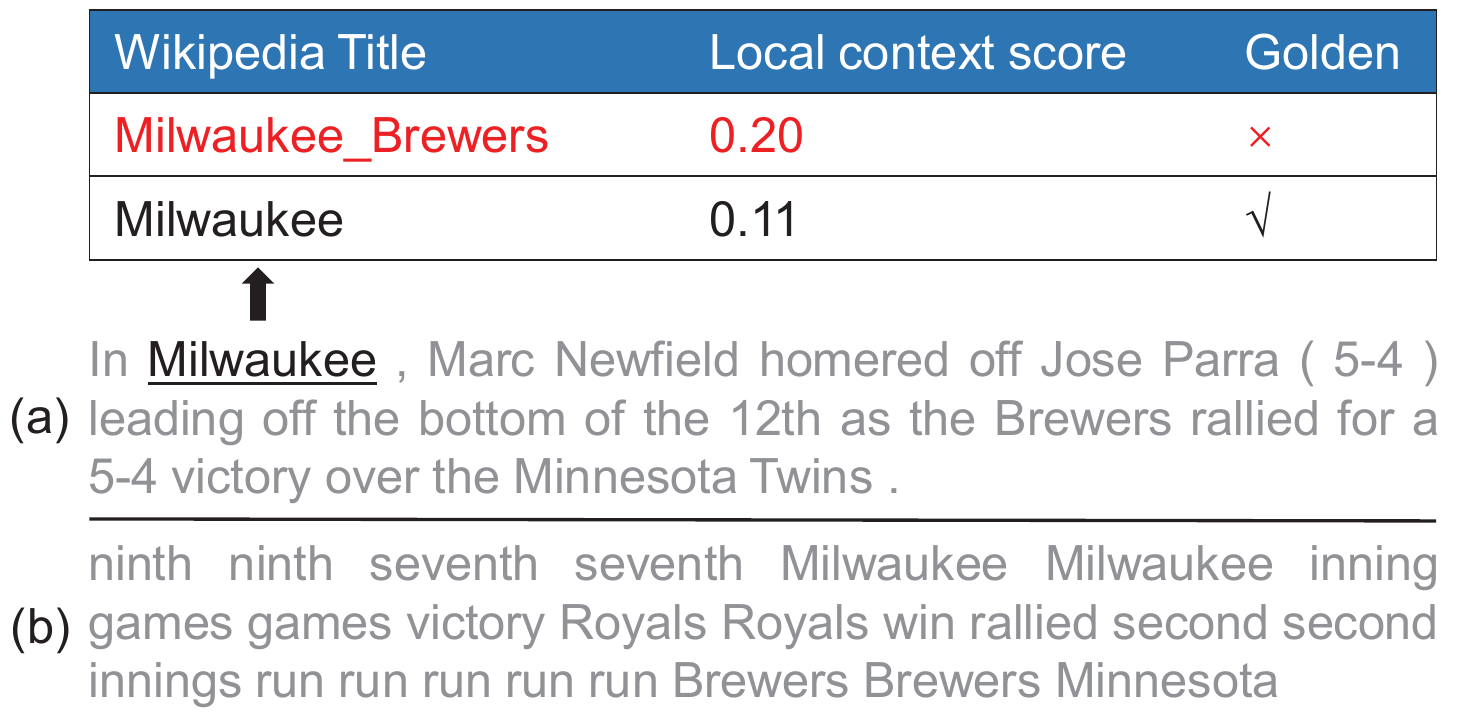}
    \caption{One error case on AIDA-CoNLL development set of the full model of \namecite{ganea2017deep}. (a) Immediate context; (b) Attended contextual words sorted by attention weights. The preposition ``In" is a strong cue predictive of the type of mention ``Milwaukee" which is not captured by the local context model. }
    \label{fig:fig1}
\end{figure}

Such state-of-the-art entity linking models~\cite{ganea2017deep,le-titov-2018-improving} employ attention-based bag-of-words context model and pre-trained entity embeddings bootstrapped from word embeddings to assess topic level context compatibility. However, the latent entity type information in the immediate context of the mention is neglected. We suspect this may sometimes cause the models link mentions to incorrect entities with incorrect type. To verify this, we conduct error analysis of the well known DeepED\footnote{https://github.com/dalab/deep-ed} model \cite{ganea2017deep} on the development set of AIDA-CoNLL \cite{hoffart-etal-2011-robust}, and found that more than half of their error cases fall into the category of type errors where the predicted entity's type is different from the golden entity's type, although some predictive contextual cue for them can be found in their local context.
As shown in Fig.~\ref{fig:fig1}, the full model of \namecite{ganea2017deep} incorrectly links the mention ``Milwaukee" to the entity \textsc{Milwaukee\_Brewers}. However, the preposition ``In" is a strong cue predictive of the type (location) of mention ``Milwaukee" which is helpful for disambiguation. The reason why the local context model of \namecite{ganea2017deep} couldn't capture such apparent cue is two folds. On one hand, the context encoding module adopts a bag-of-words encoding scheme which is position agnostic. As shown in Fig.~\ref{fig:fig1}(b), the attention mechanism is helpful for selecting predictive words (e.g. ``Milwaukee", ``games" etc.), but does not capture the pattern that the previous word ``In" of the mention ``Milwaukee" which very likely refers to an entity with location type. 
On the other hand, the pre-trained entity embedding of \namecite{ganea2017deep} is not very sensitive to entity types. 
For example, as shown in Table~\ref{table:entity_nearest}, when we query the most similar entities with the entity \textsc{Steve\_Jobs}, the top one returned entity is \textsc{Apple\_Inc.}, which is a different type but releated at topic level. So it is natural for the model of \namecite{ganea2017deep} to make type errors when it is trained to fit such entity embeddings. % it == the model

As is argued in \cite{zhou-etal-2018-zero}, ``context consistency is a strong proxy for type compatibility". Based on this claim, the mention's immediate context is a proxy of its type. For example, we consider the following context from Wikipedia linking to the entity \textsc{Apple} in which the mention is replaced with the [MASK] token. 
\begin{quote}
{\em Fruits that tend to be more popular in this area are} [MASK] {\em , pears , and berries .}
\end{quote}
By reading the context surrounding the [MASK] token, we can easily determine that the entities fitting this context should be a kind of fruit. 

In this paper, we propose to inject latent entity type information into the entity embeddings by modeling the immediate context surrounding the mention. Specifically, we apply pre-trained BERT \cite{devlin-etal-2019-bert} to represent the entity context and build a shared entity representation by aggregating all the entity contexts linking to the same entity via average pooling. Pre-trained BERT models naturally fit our purpose to represent the entity context surrounding the [MASK] token as it is trained with masked language model objective. 
What's more, we integrate a BERT-based entity similarity feature into the local model of \namecite{ganea2017deep} to better capture entity type information. This can leverage both the pre-trained entity embeddings from BERT and the domain adaption capability of BERT via fine-tuning. 

We conduct entity linking experiments on standard benchmark datasets: AIDA-CoNLL and five out-domain test sets. Our model achieves an absolute improvement of 1.32\% F1 on AIDA-CoNLL test set and average 0.80\% F1 on five out-domain test sets over five different runs. In addition, we conduct detailed experiment analysis on AIDA-CoNLL development set which shows our proposed model can reduce 67.03\% type errors of the state-of-the-art model \cite{ganea2017deep} and more than 90\% of the remaining type error cases are due to over estimation of prior and global modeling problem which we leave as the further work.

Our contributions can be summarized as follows.
\begin{itemize}
    \item We show current state-of-the-art (SOTA) neural entity linking models based on attention-based bag-of-words context model often produce type errors and analyze the possible causes. 
    \item We propose a novel entity embedding method based on pre-trained BERT to better capture latent entity type information. 
    \item We integrate a BERT-based entity similarity into the local model of a SOTA model \cite{ganea2017deep}. % omit score
    \item We verify the effectiveness of our model on standard benchmark datasets and achieve significant improvement over the baseline. And the detailed experiment analysis demonstrates that our method truly corrects most of the type errors produced by the baseline.
\end{itemize}

\section{Background}
\subsection{Entity Linking Problem}
Formally, given a document $D$ consisting of a list of entity mentions $m_1,...,m_n$. The goal of an entity linking system is to assign each $m_i$ a KB entity $e_i$ or predict that no corresponding entity in the KB (i.e., $e_i$=NIL).

Due to potentially very large entity space (e.g. Wikipedia has more than 4 million entities), standard entity linking is often divided into two stages: {\em candidate generation} which chooses potential candidates $C_i=(e_{i1},...,e_{il_i})$ using a heuristic and {\em entity disambiguation} which learns to select the best entity from the candidates using a statistical model. In this work, we focus on the second stage {\em entity disambiguation}. As for entity disambiguation, two different kinds of information can be leveraged: local context compatibility and document-level global coherence which respectively corresponds to the local model and the global model. Next, we introduce the general formulation of entity linking problem with a focus on the well known DeepED model~\cite{ganea2017deep}.

\subsubsection{General Formulation}
An entity linking model integrating both local and global features can be formulated as a conditional random field. Formally, we can define a scoring function $g$ to evaluate the entity assignment $e_1,...,e_n$ to mentions $m_1,...,m_n$ in a document $D$.
\begin{align}
    g(e_1,...,e_n|D)=\sum_{i=1}^{n} \Psi(e_i|D) + \sum_{j\neq{i}} \Phi(e_i,e_j|D)
    \label{eq:general_model}
\end{align}
where the first term scores how well an entity fits its local context and the second one measures the global coherence.

\subsubsection{Local Model}
Following \namecite{ganea2017deep}, we instantiate the local model as an attention model based on pre-trained word and entity embeddings. Specifically, for each mention $m_i$, a pruned candidate set $C_i=(e_{i1},...,e_{il_i})$ is identified in the candidate generation stage. We compute a local context score for each $e \in C_i$ based on the K-word (in practice, K is set to 100 and stop words are removed.) local context $c=\{w_1,...,w_K\}$ surrounding $m_i$. 
\begin{align}
\Psi_{\rm long}(e,c) =  \x_{e}^\top \mat B \, h(c)
\label{eq:psi}
\end{align}
where $\mat B$ is a learnable diagonal matrix, $\x_{e}$ is the embeddings of entity $e$, $h(c)$ applies a hard attention mechanism to context words in $c$ to obtain the representation of the context.

Besides, \namecite{ganea2017deep} combined this context score with the prior $\hat{p}(e|m)$ (computed by mixing mention-entity hyperlink count statistics from Wikipedia, a large Web corpus and YAGO.\footnote{See \namecite{ganea2017deep} for more details.}) using a two-layer feed-forward neural network in the local model. 
\begin{align}
\Psi(e, m, c) = f(\Psi_{\rm long}(e, c), \log \hat{p}(e|m))
\label{eq:f}
\end{align}
\subsubsection{Global Model}
The second term in Equation \ref{eq:general_model} is given by:
\begin{align}
\Phi(e,e') = \frac{2}{n-1} \,\, \x_{e}^{\top}\mat C\, \x_{e'}
\end{align}
where $\mat C$ is a diagonal matrix. The model defined by Equation \ref{eq:general_model} is a fully-connected pairwise conditional random field. Exact maximum-a-posteriori inference on this CRF, needed both at training and testing phrase, is NP-hard \cite{wainwright2008graphical}. So they used max-product loopy belief propagation (LBP) to estimate the max-marginal probability 
\begin{align}
\hat{g}_i(e|D) \approx \max_{\substack{e_1,...,e_{i-1}\\e_{i+1},...,e_n}} g(e_1,...,e_n|D)  
\end{align}
for each mention $m_i$. 
The final score for $m_i$ is given by: 
\begin{align}
	\rho_i(e) = f'(\hat{g}_i(e|D), \hat{p}(e|m_i))
\end{align}
where $f'$ is another two-layer neural network and $\hat{p}(e|m_i)$ is the prior feature.

\section{Related Work}
    Our work focuses on improving entity linking by capturing latent entity type information with BERT. Specifically, our work related to previous approaches in three aspects.
    \subsubsection{Entity Embedding}
    The entity linking task is essentially a zero-shot task where the answer of test cases may not exist in the training data.\footnote{Only 58.6\% answers of test cases in AIDA-CoNLL dataset are existent in the training data.} So we need to build a shared entity embedding space for all entities which allows neural entity linking models to generalize to both seen and unseen entities during test time. 
    Based on the distributional hypothesis \cite{harris1954distributional}, an entity is characterized by its contexts. Different methods to characterize an entity's context result in different information its entity embedding can capture. Previous work \cite{yamada2016joint,ganea2017deep} on learning entity representation are mostly extensions of the embedding methods proposed by \cite{mikolov2013distributed}. An entity's context is a bag-of-words representation which mainly captures topic level entity relatedness rather than entity type relatedness. 
    In contrast, we propose a simple method to build entity embeddings directly from pre-trained BERT \cite{devlin-etal-2019-bert} which can better capture entity type information. 

    \subsubsection{Type Information}
    Previous work attempt to integrate type information into the entity linking task mostly by jointly modeling named entity recognition and entity linking. Specifically, a line of work \cite{durrett-klein-2014-joint,luo2015joint,nguyen2016j} jointly model entity linking and named entity recognition to capture the mutual dependency between them using structured CRF. These methods mainly differ in the design of hand-engineered features. Recently, \namecite{martins-etal-2019-joint} perform multi-task learning using learned features by extending Stack-LSTM \cite{dyer-etal-2015-transition}. However, all of these work rely on extensive annotation of the type of mentions which are difficult to obtain on most of the entity linking datasets. In contrast, based on the assumption that ``context consistency is a strong proxy for type compatibility" from \namecite{zhou-etal-2018-zero}, we propose to model a mention's immediate context using BERT \cite{devlin-etal-2019-bert} to capture its contextual latent entity type information.
    \subsubsection{Applications of BERT}
    Since the advent of the well-known BERT models \cite{devlin-etal-2019-bert}, it has been applied successfully to and has achieved state-of-the-art performance on many NLP tasks. 
    The main challenges which the entity linking task has over other tasks e.g. sentence classification, named entity recognition, where BERT has been applied are: (1) a very large label space, i.e. every mention has many target entities and (2) the zero-shot nature of the entity linking task. Training label embeddings from a small labeled dataset could not generalize to cover unseen entities in test time. To tackle this problem, we introduce a novel method to build entity embeddings from BERT by modeling the immediate context of an entity.

\section{Model}
Our model consists of two phrases: (1) Build entity embeddings from BERT (2) Add a BERT-based entity similarity component to the local model. Next we will describe each phrase in the following sections.

\subsection{Entity Embeddings from BERT}
Given lists of mention context\footnote{A mention context of an entity is the surrounding text of an anchor text, i.e. mention, pointing to the entity page in Wikipedia.} $\{c_{i1}, c_{i2}, ..., c_{iN}\}$ in Wikipedia for every entity $e_i \in \mathcal{E}$, we build the entity embeddings map $\B : \mathcal{E} \rightarrow \mathbb{R}^d$. Here, the anchor context $c_{ij} = ({\rm lctx}_{ij}, {\rm m}_{ij}, {\rm rctx}_{ij})$ where ${\rm m}_{ij}$ is the mention, ${\rm lctx}_{ij}$ is the left context and ${\rm rctx}_{ij}$ is the right context. 
\subsubsection{Context Representation}
A mention's immediate context is a proxy for its type. Here, the mention's immediate context is a sequence of tokens where the mention ${\rm m}_{ij}$ is replaced with a single [MASK] token. Then, we represent the immediate entity context by extracting the upper most layer representation of pre-trained BERT \cite{devlin-etal-2019-bert} corresponding to the [MASK] token. 
\begin{align}
\mathbf{c}_{ij} = {\rm BERT}(\{{\rm lctx}_{ij}, {\rm [MASK]}, {\rm rctx}_{ij}\}) \label{context_rep}
\end{align}

\subsubsection{Entity Representation}
For each entity $e_i \in \mathcal{E}$, we randomly sample at most N anchor contexts $\{c_{i1}, c_{i2}, ..., c_{iN}\}$ from Wikipedia. Then the entity representation of $e_i$ is computed by aggregating all the context representation $\{\mathbf{c}_{i1}, \mathbf{c}_{i2}, ..., \mathbf{c}_{iN}\}$ via average pooling. 
\begin{align}
\B_{e_i} = \frac{1}{N} \sum_{j=1}^{N} \mathbf{c}_{ij}
\end{align}
As will be shown in the analysis section, the entity embeddings from BERT better capture entity type information than those from \namecite{ganea2017deep}.

\subsection{BERT-based Entity Similarity}
The local context model of \namecite{ganea2017deep} mainly captures the topic level entity relatedness information based on a long range bag-of-words context. To capture latent entity type information, we design a BERT-based entity similarity score $\Psi_{\rm BERT}(e,c)$. Specifically, given a short range (the immediate context where the mention ${\rm m}$ lies) context $c = ({\rm lctx}, {\rm m}, {\rm rctx})$, we firstly encode $c$ using the same method defined by Equation \ref{context_rep}.
\begin{align}
\mathbf{c} = {\rm BERT}(\{{\rm lctx}, {\rm [MASK]}, {\rm rctx}\}) \label{eqa:bert_c}
\end{align}
Then we define the BERT-based entity similarity as the cosine similarity\footnote{We also investigated calculating the similarity using a parameterized formula by adding a diagonal matrix between them, but found no significant improvements over Eq.~\ref{eqa:sim_cosine}.} between the context representation $\mathbf{c}$ and the entity representation $\B_{e}$.
\begin{align}
\Psi_{\rm BERT}(e,c) = {\rm cosine}(\B_{e}, \mathbf{c})    
\label{eqa:sim_cosine}
\end{align}
Finally, as for the local disambiguation model, we integrate the BERT-based entity similarity $\Psi_{\rm BERT}(e,c)$ with the local context score $\Psi_{\rm long}(e,c)$ (defined in Equation~\ref{eq:psi}) and the prior $\hat{p}(e|m_i)$ with two fully connected layers of 100 hidden units and ReLU non-linearities following the same feature composition methods as \namecite{ganea2017deep}.
\begin{align}
\Psi_{\rm local}(e, m, c) &= f(\Psi_{\rm long}(e, c), \Psi_{\rm BERT}(e, c), \\ \nonumber
&\log \hat{p}(e|m))
\end{align}
As for the global disambiguation model, we firstly define the local context score $\Psi_{\rm localctx}(e, c)$ by combining $\Psi_{\rm long}(e,c)$ and $\Psi_{\rm BERT}(e,c)$.\footnote{Following \namecite{ganea2017deep}, we did not integrate the prior score $\hat{p}(e|m)$ into the local scoring module of the global disambiguation model.}
\begin{align}
\Psi_{\rm localctx}(e, c) &= f(\Psi_{\rm long}(e, c), \Psi_{\rm BERT}(e, c))
\end{align}
Then we adopt exactly the same global model as \namecite{ganea2017deep} which is already introduced in the Background section. Specifically, we adopt loopy belief propagation (LBP) to estimate the max-marginal probability $\hat{g}_i(e|D)$ and then combine it with the prior $\hat{p}(e|m_i)$ using a two-layer neural network to get the final score $\rho_i(e)$ for $m_i$.     
    \begin{align}
        \Phi(e,e') &= \frac{2}{n-1} \,\, \x_{e}^{\top}\mat C\, \x_{e'} \\
        \hat{g}_i(e|D) &\approx \max_{\substack{e_1,...,e_{i-1}\\e_{i+1},...,e_n}} g(e_1,...,e_n|D)   \\
        \rho_i(e) &= f'(\hat{g}_i(e|D), \hat{p}(e|m_i))
    \end{align}
\subsection{Model Training}
We minimize the following max-margin ranking loss:
\begin{align}
    L(\theta)&=\sum_{D \in \mathcal{D}} \sum_{m_i \in D} \sum_{e \in C_i} h(m_i, e) \label{eqa:loss} + \lambda ||\alpha||_2^2\\ 
    h(m_i,e)&=\max\big(0, \gamma - s(e_i^*) + s(e)\big) \nonumber \\ 
    s(e) &= 
    \begin{cases}
    \Psi_{\rm local}(e, m, c)
      & \text{if local model only}\\
     \rho_i(e) & \text{local \& global model}
    \end{cases} \nonumber
\end{align}
In order to discourage the model from biasing toward a particular feature, we add a L2 regularization term  ($\lambda ||\alpha||_2^2$) w.r.t parameters $\mathbf{\alpha}$ in feature composition function $f$ to the loss function in Equation \ref{eqa:loss}, where $\lambda$ is set $10^{-7}$.

\section{Experiments}
\subsection{Datasets}
In order to verify the effectiveness of our model, we conduct experiments on standard benchmark datasets considering both in-domain and out-domain settings. For in-domain setting, we use AIDA-CoNLL dataset~\cite{hoffart-etal-2011-robust} for training, validation and testing. 
For out-domain setting, we evaluate the model trained with AIDA-CoNLL on five popular out-domain test sets: MSNBC, AQUAINT, ACE 2004 datasets cleaned and updated by \namecite{guo2018robust} and WNED-CWEB (CWEB), WNED-WIKI (WIKI) automatically extracted from ClueWeb and Wikipedia \cite{guo2018robust}. 
Following previous work \cite{ganea2017deep}, we only consider in-KB mentions. Besides, our candidate generation strategy follows that of \namecite{ganea2017deep} to make our results comparable.

\subsection{Setup}
The main goal of this work is to introduce a BERT-based entity similarity to capture latent entity type information which is supplementary to existing SOTA local context model~\cite{ganea2017deep}. So we evaluate the performance when integrating the BERT-based entity similarity into the local context model of \namecite{ganea2017deep}. We also evaluate our model with or without global modeling method of \namecite{ganea2017deep}. 
In addition, we further compare our methods with other state-of-the-art models \cite{yamada2016joint,le-titov-2018-improving}. 
To verify the contribution of our proposed BERT-based entity embeddings, we also compare with a straightforward baseline which directly replaces the encoder of \namecite{ganea2017deep} utilizing pre-trained BERT. To do so, we introduce a $768 \times 300$ dimensional matrix $\mat W$ which projects BERT-based context representation $\mathbf{c}$ into \namecite{ganea2017deep}'s entity embeddings space when calculating the similarity score.
\subsection{Hyper-parameter Setting}
The resources (word and entity embeddings) used to train the local context model of \namecite{ganea2017deep} are obtained from DeepED\footnote{https://github.com/dalab/deep-ed/}. For each entity, we randomly sample at most 100 anchor contexts from Wikipedia\footnote{Using the same Wikipedia dump (Feb 2014) as the one which \namecite{ganea2017deep} used to train their entity embeddings.} to build the entity representation from BERT. We discard any articles appearing in WIKI dataset when building the entity representation from BERT. 
We take the anchor context as the surrounding sentence where the mention lies and replace the mention with a single [MASK] token. 
Each context is truncated to 128 tokens after WordPiece tokenization.
We use the PyTorch implementation of pre-trained BERT models\footnote{https://github.com/huggingface/pytorch-pretrained-BERT} and choose the BERT-base-cased version. We adopt the Adam~\cite{kingma2014adam} implemented by BERT with $\beta_1=0.9$, $\beta_2=0.999$. Empirically, we found that it is helpful to set parameters in BERT a small initial learning rate and not BERT related parameters a larger initial learning rate to avoid the whole model biasing toward the BERT feature and disregarding other model components. 
In our experiments, pre-trained BERT model is fine-tuned with initial learning rate $10^{-5}$ whereas not BERT related parameters are trained with $10^{-3}$. 
Similar learning rate usage can be found in the recent work by \cite{hwang2019comprehensive}. 
Similar to \namecite{ganea2017deep}, all the entity embeddings are fixed during fine-tuning. We randomly initialize the not BERT related parameters using Gaussian distribution $\mathcal{N}(0.0, 0.02)$ and the bias term is zeroed. 

Note that all the hyper-parameters used in the local context and global model of \namecite{ganea2017deep} were set to the same values as theirs for direct comparison purpose. Detailed hyper-parameters setting is described in the appendices. Our model is trained with 4 NVIDIA Tesla P100 GPUs. We run each of our model five times with different random seeds, and the performance is reported in the form of average $\pm$ standard deviation.

\begin{table}[!t]
    	\small
        \centering
	    \scalebox{0.90}{
        \begin{tabular}{lc}
            \hline
        	Methods & AIDA-B \\
        	\hline
        	\emph{Local models}                                \\
        	prior $\hat{p}(e|m)$ & 71.9                         \\
        	\namecite{lazic-etal-2015-plato} & 86.4                        \\
            \namecite{globerson2016collective} & 87.9               \\
            \namecite{yamada2016joint} & 87.2                       \\
            \namecite{ganea2017deep} &   88.8                 \\
            \namecite{ganea2017deep} (reproduce) & $88.75 \pm 0.30$ \\
            BERT-Entity-Sim (local) & $\mathbf{90.06} \pm 0.22$  \\
            \hline
            \emph{Local \& Global models}                               \\
            \namecite{huang2015leveraging} & 86.6 \\
            \namecite{ganea2016probabilistic} & 87.6\\
            \namecite{chisholm-hachey-2015-entity} & 88.7 \\
            \namecite{guo2018robust} & 89.0 \\
            \namecite{globerson2016collective} & 91.0 \\
            \namecite{yamada2016joint} & 91.5 \\
            \namecite{ganea2017deep} & $92.22 \pm 0.14$ \\
            \namecite{le-titov-2018-improving} & $93.07 \pm 0.27$ \\
            \hline
            \multicolumn{2}{l}{\emph{Explicit entity type injection models in Analysis section}}\\
            Oracle type (Ultra-fine) & $95.38 \pm 0.07 $\\
            Oracle type (FIGER) & $96.35 \pm 0.14$ \\
            Predict type (Ultra-fine) & $91.35 \pm 0.18$\\
            Predict type (ZOE) & $91.42 \pm 0.09$\\
            \hline
            BERT+G\&H's embeddings & $91.00 \pm 0.72$\\
            BERT-Entity-Sim (local \& global) & $\mathbf{93.54} \pm 0.12$ \\
            \hline
        \end{tabular}}
        \caption{F1 scores on AIDA-B (test set).}
        \label{tab:in-domain test}
    \end{table}

\subsection{Results}
    \begin{table*}[!t]
	\small
    \centering
    \scalebox{0.90}{
    \begin{tabular}{lccccc|c}
        \hline
    	Methods & MSNBC & AQUAINT & ACE2004 & CWEB & WIKI & Avg \\
        \hline
        prior $\hat{p}(e|m)$          & 89.3 & 83.2 & 84.4 & 69.8 & 64.2 & 78.18 \\
        \hline
        \namecite{milne2008learning} & 78 & 85 & 81 & 64.1 & 81.7 & 77.96 \\
        \namecite{hoffart-etal-2011-robust} & 79 & 56 & 80 & 58.6 & 63 & 67.32 \\
        \namecite{ratinov-etal-2011-local} & 75 & 83 & 82 & 56.2 & 67.2 & 72.68 \\
        \namecite{cheng-roth-2013-relational} & 90 & $\mathbf{90}$ & 86 & 67.5 & 73.4 & 81.38 \\
        \namecite{guo2018robust} & 92 & 87 & 88 & 77 & $\mathbf{84.5}$ & 85.70 \\
        \namecite{ganea2017deep} & 93.7 $\pm$ 0.1 & 88.5 $\pm$ 0.4 & 88.5 $\pm$ 0.3 & $\mathbf{77.9}$ $\pm$ 0.1 & 77.5 $\pm$ 0.1 & 85.22 \\
        \namecite{le-titov-2018-improving} & $\mathbf{93.9}$ $\pm$ 0.2 & 88.3 $\pm$ 0.6 & $\mathbf{89.9}$ $\pm$ 0.8 & 77.5 $\pm$ 0.1 & 78.0 $\pm$ 0.1 & 85.51 \\
        \hline
        \multicolumn{3}{l}{\emph{Explicit entity type injection models in Analysis section}}\\
        Oracle type (Ultra-fine) & 96.8 $\pm$ 0.1 & 93.7 $\pm$ 0.2 & 92.0 $\pm$ 0.2 & 85.8 $\pm$ 0.1 & 84.0 $\pm$ 0.1 & 90.46 \\
        Oracle type (FIGER) & 97.1 $\pm$ 0.1 & 92.3 $\pm$ 0.2 & 93.1 $\pm$ 0.2 & 84.3 $\pm$ 0.2 & 84.4 $\pm$ 0.1 & 90.24 \\
        Predict type (Ultra-fine) & 93.4 $\pm$ 0.2 & 89.7 $\pm$ 0.2 & 89.1 $\pm$ 0.5 & 77.8 $\pm$ 0.1 & 76.8 $\pm$ 0.2 & 85.36\\
        Predict type (ZOE) & 93.2 $\pm$ 0.2 & 89.5 $\pm$ 0.2 & 89.2 $\pm$ 0.4 & 77.6 $\pm$ 0.1 & 77.0 $\pm$ 0.1 & 85.30 \\
        \hline
        BERT+G\&H's embeddings & 93.3 $\pm$ 0.4 & 89.1 $\pm$ 0.6 & 88.1 $\pm$ 0.6 & 75.7 $\pm$ 0.5 & 76.3 $\pm$ 0.5 & 84.50 \\
        BERT-Entity-Sim (local \& global) & 93.4 $\pm$ 0.1 & \underline{89.8} $\pm$ 0.4 & 88.9 $\pm$ 0.7 & $\mathbf{77.9}$ $\pm$ 0.4 & \underline{80.1} $\pm$ 0.4 & $\mathbf{86.02}$ \\
        \hline
    \end{tabular}}
    \caption{F1 scores on five out-domain test sets. Underlined scores denote the corresponding model outperforms the baseline.}
    \label{tab:out-domain test}
    \end{table*}
    
    \begin{table*}[!t]
	\small
    \centering
    \scalebox{0.90}{
    \begin{tabular}{l|ccc|ccc|ccc}
    \hline
     & \multicolumn{3}{c|}{\figer}  & \multicolumn{3}{c|}{\bbn} & \multicolumn{3}{c}{\ontonotesFine}                          \\
        \hline
    Entity Embedding & \fOneMi & \fOneMa & \accuracy & \fOneMi & \fOneMa & \accuracy & \fOneMi & \fOneMa & \accuracy \\
        \hline
    \namecite{ganea2017deep} & 80.38 & 82.65 & 53.30 & 80.87 & 84.10 & 69.34 & 81.41 & 83.54 & 57.54 \\
    BERT based Entity Embedding & 88.69 & 90.98 & 69.07 & 91.30 & 93.35 & 85.36 & 90.74 & 92.52 & 73.94 \\
    \hline
    \end{tabular}}
    \caption{Results of type classification task on three typing systems: \figer, \bbn, \ontonotesFine} 
        \label{tab:probe_task}
    \end{table*}
    
Table~\ref{tab:in-domain test} shows the micro F1 scores on in-domain AIDA-B dataset of the SOTA methods and ours, which all use Wikipedia and YAGO mention-entity index. The models are divided into two groups: {\em local models} and {\em local \& global models}. As we can see, our proposed model, BERT-Entity-Sim, outperforms all previous methods. Our local model achieves a 1.31 improvement in terms of F1 over its corresponding baseline \cite{ganea2017deep}, yielding a very competitive local model with an average 90.06 F1 score even surpassing the performance of four local \& global models. Equipped with the global modeling method of \namecite{ganea2017deep}, the performance of our model further increase to 93.54 with an average 1.32 improvement in terms of F1 over \namecite{ganea2017deep}. In addition, our method outperforms \namecite{le-titov-2018-improving} model by 0.47 point. 
The model of \namecite{le-titov-2018-improving} is a multi-relational extension of \namecite{ganea2017deep}'s global modeling method while keeps exactly the same local context model. 
Our better local context model should be orthogonal with them and has potential more applications on short texts (e.g. tweets) where global modeling has little benefits.
Moreover, BERT+G\&H’s embeddings performs significantly worse than the baseline \cite{ganea2017deep} and our proposed BERT-Entity-Sim model. The reason is that BERT-based context representation space and \citeauthor{ganea2017deep}’s entity embeddings space are heterogeneous. \citeauthor{ganea2017deep}’s entity embeddings are bootstrapped from word embeddings which mainly capture topic level entity relatedness, while BERT-based context representation is derived from BERT which naturally captures type information. The non-parallel information in both context and entity sides makes it difficult to learn the alignment parameter $\mat W$ and results in the poor generalization performance.

To evaluate the robustness of our model, Table~\ref{tab:out-domain test} shows the performance of our method and SOTA methods on five out-domain test sets. 
On average, our proposed model (BERT-Entity-Sim) outperforms the local \& global version of \namecite{ganea2017deep,le-titov-2018-improving} by an average 0.80 and 0.51 on F1. 

\subsection{Analysis}

    \begin{table}[ht]
    	\small
        \centering
        \scalebox{0.85}{
        \begin{tabular}{lcc}
            \hline
        	Error Type & \# Cases & Percentage (\%) \\
        	\hline
        	{\em Due to prior} & 41 & 67.21 \\
        	{\em Due to global} & 14 & 22.95 \\
        	{\em Due to local context} & 6 & 9.84 \\
        	\hline
        \end{tabular}}
        \caption{Remaining type error cases categorization}
        \label{tab:remain_error_category}
    \end{table}
    
    \begin{table}[!t]
	    \small
        \centering
        \scalebox{0.90}{
        \begin{tabular}{cccc}
        \hline
        Typing System & \fOneMi & \fOneMa & \accuracy \\
        \hline
        \namecite{choi-etal-2018-ultra} & 26.52\% & 26.60\% & 0.36\%\\
        \namecite{zhou-etal-2018-zero} & 66.12\% & 67.98\% & 46.08\% \\
        \hline
        \end{tabular}}
        \caption{Performance of two state-of-the-art fine grained entity typing systems on AIDA-CoNLL development set}
        \label{tab:type_sys_performance}
    \end{table}

We conduct experiment analysis to answer the following questions:
\begin{itemize}
    \item Do the entity embeddings from BERT better capture latent entity type information than that of \namecite{ganea2017deep}?
    \item Does the proposed model correct the type errors in the baseline \cite{ganea2017deep}?
    \item Can straightforward integration of state-of-the-art fine grained entity typing systems improve entity linking performance?
    \item Can better global model further boost the performance of the proposed model?
\end{itemize}

    \begin{table*}[!t]
    	\small
        \centering
        \scalebox{0.90}{
        \begin{tabular}{l||c||ccccc|c}
            \hline
        	Methods & AIDA-B & MSNBC & AQUAINT & ACE2004 & CWEB & WIKI & Avg \\
            \hline
            \namecite{ganea2017deep} & 92.22 $\pm$ 0.14 & 93.7 $\pm$ 0.1 & 88.5 $\pm$ 0.4 & 88.5 $\pm$ 0.3 & 77.9 $\pm$ 0.1 & 77.5 $\pm$ 0.1 & 85.22 \\
            \namecite{le-titov-2018-improving} & 93.07 $\pm$ 0.27 & 93.9 $\pm$ 0.2 & 88.3 $\pm$ 0.6 & 89.9 $\pm$ 0.8 & 77.5 $\pm$ 0.1 & 78.0 $\pm$ 0.1 & 85.51 \\
            \namecite{yang2019learning} & $\mathbf{94.64}$ $\pm$ 0.20 & $\mathbf{94.6}$ $\pm$ 0.2 & 87.4 $\pm$ 0.5 & 89.4 $\pm$ 0.4 & 73.5 $\pm$ 0.1 & 78.2 $\pm$ 0.1 & 84.62 \\
            \hline
            BERT-Entity-Sim (local \& global) & 93.54 $\pm$ 0.12 & 93.4 $\pm$ 0.1 & $\mathbf{89.8}$ $\pm$ 0.4 & 88.9 $\pm$ 0.7 & 77.9 $\pm$ 0.4 & 80.1 $\pm$ 0.4 & 86.02 \\
            BERT-Entity-Sim (local \& DCA global) & 93.66 $\pm$ 0.17 & 94.5 $\pm$ 0.3 & 89.1 $\pm$ 0.3 & $\mathbf{90.8}$ $\pm$ 0.4 & $\mathbf{78.2}$ $\pm$ 0.2 & $\mathbf{81.0}$ $\pm$ 0.3 & $\mathbf{86.72}$ \\
            \hline
        \end{tabular}}
        \caption{F1 scores of BERT-Entity-Sim equipped with the DCA global model \cite{yang2019learning} on six test sets.}
        \label{tab:dca-test}
        \end{table*}
        
\subsubsection{Effectiveness of BERT-based and Ganea \& Hofmann (2017) Entity Embedding in Entity Type Prediction} In order to verify our claim that the entity embeddings from BERT better capture entity type information than those from \namecite{ganea2017deep}, we carry out an entity type prediction task based on its entity embedding. Specifically, we randomly sample 100K entities from Wikipedia, and randomly split them into training set (80K), development set (10K) and test set (10K). For each entity, we obtain its entity types from three typing systems: \figer ~\cite{ling2012fine}, \bbn ~\cite{weischedel2005bbn} and \ontonotesFine ~\cite{gillick2014context} via the entity type mapping provided by \namecite{zhou-etal-2018-zero}. The entity type prediction model is a simple linear classification model\footnote{Due to space limitation, we put the detailed description of this model and training hyper-parameters in the Appendices.} using the entity embedding of an entity as features; limiting its capacity enables us
to focus on whether type information can be easily extracted from the entity embeddings.
We evaluate the model using standard entity typing metrics: Strict Accuracy (\accuracy), Micro F1 (\fOneMi) and Macro F1 (\fOneMa). 

As shown in Table~\ref{tab:probe_task}, our proposed entity embedding from BERT significantly outperforms the entity embedding proposed by \namecite{ganea2017deep} on three typing systems \figer, ~\bbn ~ and \ontonotesFine. Specifically, our method improves over the baseline with an absolute 8.31, 10.43 and 9.33 \fOneMi ~ point than the baseline on three typing systems respectively. This demonstrates that our proposed entity embeddings from BERT indeed capture better latent entity type information than \namecite{ganea2017deep}.

\subsubsection{Type Errors Correction}

As we have mentioned in the introduction section, more than half of the baseline model's errors on the AIDA-A dataset are type errors. Type errors are error cases\footnote{We discard the error cases due to candidate generation problem (i.e., gold entities that do not appear in mentions' candidate list) which cover 2.98\% mentions of AIDA-A dataset.} where (1) the predicted entity's type is different from the golden entity's type; (2) contextual cue predictive of the type of the mention exists; (3) errors are not due to annotation errors. By doing so, we collect 185 type error cases which cover 57.45\% of all (322) error cases. This indicates that \namecite{ganea2017deep} produces many type errors due to its inability to consider the entity type information in mention context. By integrating the BERT-based entity similarity, our proposed model can correct 124 out of 185 (67.03\%) type error cases of the baseline model which demonstrates that we correct more than two third of the type errors produced by the baseline. We have further examined and categorized the remaining 61 type error cases into three categories: (i) {\em Due to prior}: golden entities with very low $\hat{p}(e|m_i)$ prior, (ii) {\em Due to global}: both the local context score and prior score support predicting the golden entity, but the overall score supports predicting incorrect entity due to global modeling, (iii) {\em Due to local context}: the local context score misleads the model predicting the wrong entity, this is potentially due to the mention context can be misleading, e.g. a document discussing cricket will favor resolving the mention ``Australian" in context ``impressed by the positive influence of {\em Australian} coach Dave Gilbert" to the entity \textsc{Australia\_national\_cricket\_team} instead of the gold entity \textsc{Australia}. 

As shown in Table \ref{tab:remain_error_category}, 67.21\% of the remaining type error cases are due to {\em prior} problem which are hard to solve in the current feature combination framework. We argue that prior should be considered as the final resort, only relying on it when the model can not make decision based on other features. Besides, there are 22.95\% remaining type errors which are due to {\em global modeling} problem which shows the limitation of the global modeling method of \namecite{ganea2017deep}. Finally, 9.84\% type error cases are due to {\em local context} problem that our BERT-based solution cannot address. We leave this to future work.

\subsubsection{Incorporating Explicit Entity Types} We have shown that our BERT-based local context model which implicitly captures entity type information and is effective in correcting two third of type error cases. It is nature to conjecture that we can also correct type errors by incorporating explicit type information into \namecite{ganea2017deep}. We investigate this approach in this section. Assuming that we have types for each mention and candidate entity, we calculate the Jaccard similarity between them and use it as a feature for local disambiguation model.
\begin{align}
    {\rm JaccardSim}(e, m, c) = \frac{|T_m \cap T_e|}{|T_m \cup T_e|}
\end{align}
where $T_m$ and $T_e$ are the type sets of the mention $m$ and candidate entity $e$ respectively.
The new local context score function considering explicit type information is defined as:  
\begin{align}
\Psi'_{\rm localctx}(e, m, c) &= f\big(\Psi_{\rm long}(e, c), {\rm JaccardSim}(e, m, c)\big) \nonumber
\end{align}
We consider both \textbf{Oracle} setting and \textbf{Predict} setting.
In the oracle setting, the mention's types are set as the golden entity's types.\footnote{In practice, this setting is unachievable due to potentially insufficient context and the imperfect entity typing system.}
As for the entity's types, we use two sources: one is the ultra-fine type sets from \namecite{choi-etal-2018-ultra} consisting of more than 10,000 ultra-fine grained types; the other one is the FIGER type sets \cite{ling2012fine} consisting of 112 fine grained types. In the predict setting, we use two state-of-the-art fine grained entity typing systems: 1) Ultra-fine \cite{choi-etal-2018-ultra} which predicts types in ultra-fine type sets; 2) ZOE \cite{zhou-etal-2018-zero} which can predict types in FIGER type sets. 

As we can see from both Table \ref{tab:in-domain test} and Table \ref{tab:out-domain test}, in the oracle setting, the best model outperforms all the state-of-the-art entity linking models by a large margin, even surpass \namecite{le-titov-2018-improving} by 3.28 F1 points on AIDA-CoNLL test set. This result shows that a better type prediction system can further improve upon the state-of-the-state entity linking systems. However, in the predict setting, the type injection models have worse performance than the baseline. The degradation might be attributed to the poor performance of the two state-of-the-art fine grained entity typing systems. To verify this, we measure the performance of the two typing systems on AIDA-CoNLL development set.\footnote{The mention's golden types are set as its linked entity's types.} As shown in Table ~\ref{tab:type_sys_performance}, the ultra-fine entity typing system \cite{choi-etal-2018-ultra} only achieves 26.52\% \fOneMi~score while the ZOE system \cite{zhou-etal-2018-zero} achieves 66.12\% \fOneMi~score\footnote{The size of type sets of ultra-fine typing system is much larger than that of ZOE.}
which are insufficient to improve state-of-the-art entity linking system with more than 92\% F1 score. 

\subsubsection{Better Global Model}
    In order to investigate whether better global model can further boost the performance of our model, we incorporate the recent proposed Dynamic Context Augmentation (DCA)\footnote{https://github.com/YoungXiyuan/DCA/} \cite{yang2019learning}. 
    DCA is a global entity linking model featuring better efficiency and effectiveness than that of \namecite{ganea2017deep} by 
    breaking the ``all-mention coherence" assumption. 
    Compared to BERT-Entity-Sim equipped with \namecite{ganea2017deep}'s global model in Table 8, BERT-Entity-Sim (local \& DCA global) model gains a further improvement of 0.12 F1 on AIDA-B and 0.70 F1 on five out-domain test sets on average.
    This shows that better global model indeed can further boost the performance of our proposed model.
    
    Notice that \namecite{yang2019learning} achieves a high performance on in-domain AIDA-CoNLL test set, but doesn't perform well on five out-domain test sets (even inferior than \namecite{ganea2017deep}). %After investigating this phenomenon
    We found that \namecite{yang2019learning} includes an explicit type similarity which is based on a typing system\footnote{It yields 95\% accuracy on AIDA-A according to their paper.} trained with AIDA-train NER annotation. This explicit type similarity feature is tailored for AIDA-CoNLL data set and doesn't achieve good generalization performance on out-domain test sets. In contrast, our BERT-Entity-Sim model capturing latent type information has potential better generalization performance with an average 2.10 F1 improvement over them.
\subsection{Case Study}
    
    \begin{table*}[!t]
    \centering
    \small
    \scalebox{0.85}{
        \begin{tabular}{clcc}
        \hline
        \textbf{Model} & \textbf{Context} & \textbf{Context Sim} & \textbf{Golden Entity} \\
        \hline
        \rowcolor{lightgrey}
        \multirow{6}{*}{Our}& In \textbf{Milwaukee} , Marc Newfield homered off Jose Parra ( 5-4 ) ... & - &  \textsc{Milwaukee} \\ 
        & (1) In \textbf{Cleveland} , Kevin Seitzer 's two-out single ... & 0.968 & \textsc{Cleveland} \\
        & (2) In \textbf{Boston} , Troy O'Leary homered off the right-field ... & 0.956 & \textsc{Boston} \\
        & (3) In \textbf{Houston} , Jeff Bagwell homered and Donne Wall ... & 0.951 & \textsc{Houston} \\
        & (4) In \textbf{Los Angeles} , Greg Gagne had a run-scoring single ... & 0.949 & \textsc{Los\_Angeles} \\
        & (5) In \textbf{Houston} , Andy Benes allowed two runs over seven innings ... & 0.947 & \textsc{Houston} \\
        \hline
        \rowcolor{lightgrey}
        \multirow{6}{*}{Baseline}& ninth ninth seventh seventh Milwaukee Milwaukee inning games games victory ... & - &  \textsc{Milwaukee} \\ 
        & (1) eighth ninth Milwaukee inning league runs victory fourth rallied earned ... & 0.941 & \textsc{Milwaukee\_Brewers} \\
        & (2) Denny runs fifth fifth allowed game fourth San win win ... & 0.940 & \textsc{Philadelphia\_Phillies} \\
        & (3) Royals league games runs game won won win Minnesota straight ... & 0.938 & \textsc{Minnesota\_Twins} \\
        & (4) Cleveland fifth games sixth inning innings Friday extra Sox month ... & 0.934 & \textsc{Cleveland\_Indians} \\
        & (5) games streak stay Reynoso runs runs second run won fourth ... & 0.926 & \textsc{Miami\_Marlins} \\
        \hline
        \end{tabular}}
    \caption{Nearest contexts for the example in Fig.~\ref{fig:fig1} in BERT's and baseline's context representation space}
    \label{tab:sim_ctx}
    \end{table*}

\begin{table*}[!t]
        \centering
        \small
        \scalebox{0.90}{
        \begin{tabular}{llll}
        \hline
        Model  & \textsc{Steve\_Jobs} & \textsc{National\_Basketball\_Association} & \textsc{Beijing} \\
        \hline
        \multirow{3}{*}{\namecite{ganea2017deep}} & \textsc{Apple\_Inc.} & \textsc{Sacramento\_Kings} & \textsc{Seoul} \\
                                                  & \textsc{Steve\_Wozniak} & \textsc{Golden\_State\_Warriors} & \textsc{Shanghai} \\
                                                  & \textsc{Bill\_Gates} & \textsc{Los\_Angeles\_Clippers} & \textsc{China} \\
        \hline
        \multirow{3}{*}{BERT based Entity Embedding}                 & \textsc{Steve\_Wozniak} & \textsc{American\_Basketball\_Association} & \textsc{Guangzhou} \\
                                                  & \textsc{Bill\_Gates} & \textsc{Women's\_National\_Basketball\_Association} & \textsc{Shanghai} \\
                                                  & \textsc{Steve\_Ballmer} & \textsc{National\_Basketball\_League\_(United\_States)} & \textsc{Nanjing} \\
        \hline
        \end{tabular}}
        \caption{Examples of nearest entities in \namecite{ganea2017deep} and BERT based entity representation space}
        \label{table:entity_nearest}
    \end{table*}
    
We demonstrate the effectiveness of our proposed model by retrieving the nearest neighbours in both the context representation space and entity representation space.
\subsubsection{Nearest Contexts}
    We follow \namecite{papernot2018deep} to retrieve training examples using nearest neighbour in the context representation space. For our model, we use the context representation $\mathbf{c}$ defined by Equation \ref{eqa:bert_c}. For the baseline model, we use the attention-based context representation $h(c)$ defined in Equation \ref{eq:psi}. This can reveal which training instances support the prediction of a model. As shown in Table \ref{tab:sim_ctx}, for the example in Figure \ref{fig:fig1}, the most similar contexts retrieved by our model's context representation are all with preposition ``In" ahead of the mention and the golden entities of them are all American cities. In contrast, the baseline's local context is a bag-of-words representation which we denote using the top 10 attended contextual words sorted by attention weights. The most similar contexts retrieved by baseline's context representation share common words like ``games", ``victory" and the golden entities of them are all baseball teams. This explains why the baseline model incorrectly links the mention ``Milwaukee" to \textsc{Milwaukee\_Brewers} while our model can link to the correct entity \textsc{Milwaukee}. 
\subsubsection{Nearest Entities}
    We also retrieve nearest entities in the embedding space of \namecite{ganea2017deep} and ours. As we can see, we query \textsc{Steve\_Jobs}, the nearest entity in \namecite{ganea2017deep} is \textsc{Apple\_Inc.} which is a different type. In contrast, all the entities retrieved by our approach share the same types like person, entrepreneur etc. Another example is when we query \textsc{National\_Basketball\_Association}, the most similar entities in \namecite{ganea2017deep} are NBA teams which are topically related, while the entities retrieved by our approach are all basketball leagues. 
\section{Conclusion}
    In this paper, we propose to improve entity linking by capturing latent entity type information with BERT. Firstly, we build entity embeddings from BERT by averaging all the context representation extracted from pre-trained BERT. Then we integrate a BERT-based entity similarity into the local model of the state-of-the-art method by \cite{ganea2017deep}. The experiment results show that our model significantly outperforms the baseline with an absolute improvement of 1.32\% F1 on in-domain AIDA-CoNLL test set and average 0.80\% F1 on five out-domain test datasets. The detailed experiment analysis shows that our method corrects most of the type errors produced by the baseline. In the future, we would like to design global modeling methods which can take advantage of the BERT architecture and investigate other ways to use the prior feature. 
    
\section{Acknowledgments}    
This work is partly funded by National Key Research and Development Program of China via grant 2018YFC0806800 and 2018YFC0832105.

\section{Appendices}

\subsection{Classification Model of Entity Prediction Task}

Given an entity $e$, we firstly retrieve its entity embedding $\mathbf{e}$, then compute the probability for each type in the typeset $T$: 
\begin{align}
    p_j^e = \sigma(\mathbf{w_j}^\top \mathbf{e} + b)
\end{align}
where $\sigma$ is the sigmoid function, $\mathbf{w_j}$ and $b$ are respectively the weight and bias parameter. For each entity $e$, it is labeled with $t^e$, a binary vector of all types where $t_j^e=1$ if the $j^{\rm th}$ type is in the set of gold types of $e$ and 0 otherwise. We optimize a multi-label binary cross entropy objective:
\begin{align}
    L_{\rm type}= -\sum_j t_j^e \log p_j^e + (1-t_j^e)\log(1-p_j^e)
\end{align}
We optimize the model with Adam with an initial learning rate of 1e-3. Each model is trained for up to 200 epoches and training stops when the performance on the development set does not improve for 6 consecutive epoches. 

\subsection{Detailed Hyper-parameters Setting (Table~\ref{tab:hyper-parameters})}
\label{app:hp}

    \begin{table}[!ht]
        \small
        \centering
         \scalebox{0.90}{
            \begin{tabular}{l|c}
            \hline
            Hyper-parameters & Value \\
            \hline
            BERT-based entity embedding dims & 768 \\
            dumping factor & 0.5 \\
            number of LBP loops & 10 \\
            batch size & 1 document \\ 
            & ($\leq 64$ mentions)\\
            $\gamma$ (margin) & 0.01\\
            epoch (local model) & 2\\
            epoch (local \& global model) & 10\\
            \hline
            \end{tabular}}
        \caption{Values of hyper-parameters.}
        \label{tab:hyper-parameters}
    \end{table}

\bibliography{aaai20}
\bibliographystyle{aaai}

\end{document}